\pdfoutput=1
\documentclass[11pt]{diazessay} 
\usepackage[dvipsnames]{xcolor}
\usepackage[colorlinks,allcolors=NavyBlue]{hyperref}
\usepackage{natbib}
\usepackage{enumitem}
  
\title{\textbf{Evaluating NLG systems} \\ {\Large\itshape A brief introduction}} 

\author{\textbf{Emiel van Miltenburg} \\ \textit{Tilburg University}} 

\date{\today} 

\begin{document}

\maketitle 


\renewcommand{\abstractname}{Summary} 

\begin{abstract}
This year the International Conference on Natural Language Generation (INLG) will feature an award for the paper with the best
evaluation. The purpose of this award is to provide an incentive for NLG
researchers to pay more attention to the way they assess the output of
their systems. This essay provides a short introduction to evaluation in
NLG, explaining key terms and distinctions.
\end{abstract}

\section*{How can I evaluate my system?}
It is hard to say in general how you should evaluate your NLG system.
Much depends on the kind of system that you are developing, and the
context in which it is being used. A first step is to get used to
commonly used terminology in the field, so that you know what
possibilities are out there.

\begin{description}[wide]
\item[Intrinsic evaluation] refers to the assessment of system output in
isolation. For example, grammaticality is a property that you can assess
using intrinsic evaluation. You could use either \emph{human evaluation}
(e.g.~grammaticality judgments) or \emph{automatic metrics} (e.g.~a
precision grammar or a grammar checker) to determine whether the output
of an NLG system is grammatical. (For an overview of different
properties that you could evaluate, see
\citealt{belz-etal-2020-disentangling}.)
Finally, you could carry out an \emph{error analysis} to determine where
the system still falls short.

\item[Extrinsic evaluation] refers to the impact that a system may have
on downstream processes. If you have a newly developed NLG system, you
could for example see whether employees become more efficient or more
productive after the system has been deployed.
\end{description}

Sometimes people also use the term ``extrinsic evaluation'' to assess
the impact of a specific module in an NLG pipeline. An \emph{intrinsic}
evaluation of an NLG pipeline module would just look at the quality of
its output, rather than the final text that is produced by the full
pipeline. You could carry out an \emph{extrinsic} evaluation by
determining the extent to which the final output improves when you
replace an existing module (e.g.~for rule-based referring expression
generation) with a newly developed module.

If you are developing multiple new modules, you could carry out
\emph{ablation tests} (systematically leaving out or replacing different
modules) to see how much each new module contributes to the system's
overall performance.

\section*{Human versus automatic evaluation}

Human evaluation is generally seen as the gold standard in NLG research,
because in the end it is essential that human readers appreciate the
output of your system. Having that said, there is great value in
reliable automatic metrics, since they are cheaper and not as
labor-intensive. Reliability is the keyword here: no matter how you
measure different properties of the generated text, we have to be able
to trust the conclusions that you draw from your observations.

\section*{What are current best practices in the field?}

It is always risky to talk about best practices, because evaluation is
so context-dependent. As the saying goes: ``a foolish consistency is the
hobgoblin of little minds'' \citep{Emerson1982-zx}. For some projects, it may be
better to deviate from existing standards. Having said that, here are
ten steps you may find useful in planning your evaluation. Some of these
steps raise fundamental questions about your research project, so it is
important to start thinking about evaluation at the start of your
project, and to not consider evaluation as an afterthought. Moreover,
good evaluations take time, which means you need to schedule enough time
to carry out a reliable evaluation.

\begin{enumerate}
\def\labelenumi{\arabic{enumi}.}
\item
  \emph{Determine the target audience.} Who are you developing your NLG
  system for? Ideally you will evaluate the performance of your system
  with a group of participants that matches the demographic properties
  of your target audience. Even if you don't carry out a human
  evaluation, it is still important to understand the application
  context because of the next point.
\item
  \emph{Get to know user needs}. If you have an applied NLG project, you
  should ideally start from an understanding of the stakeholders' needs.
  That means talking to the people who will use your system, getting to
  know what they want to do with your system and what properties are
  important for them. Then you can develop an evaluation protocol that
  is in line with user needs.
\item
  \emph{Identify relevant work.} Search for relevant literature,
  identifying models you want to compare to and metrics that are
  commonly used. Determine if those metrics make sense given your
  project. Consider reproducing relevant results, to be able to carry
  out any comparisons yourself.
\item
  \emph{Determine your goals and expectations}. Based on the earlier
  steps, formulate a relevant research question. Think about possible
  outcomes of your project, which outcome is more likely, and why you
  expect this to be the case. Also consider how different outcomes of
  your evaluation should be interpreted. Try to be as precise as
  possible. Does it make sense to form hypotheses about your
  experiments, and to motivate them on the basis of earlier literature?
  This helps with theory building in NLG. As a part of this process, you should:
  \begin{itemize}

  \item
    \emph{Identify key independent variables.} There are many factors
    that could influence the characteristics of the output text.
    Different sets of inputs lead to different kinds of outputs, and
    different system properties affect the way the output text looks.
    Your job is to identify the main variables of interest.
  \item
    \emph{Determine key dependent variables}. There are many different
    properties of NLG output that you could assess, for example:
    grammaticality, fluency, completeness, naturalness, appropriateness,
    and the list goes on. You don't have space to cover them all
    (although you could provide an additional extensive evaluation in
    the appendix), and some properties are probably more relevant than
    others for the purpose of your project. Clearly define the
    constructs of interest before you start thinking about how to
    operationalise those constructs. Use those definitions to critically
    assess the metrics you are planning to use.
  \end{itemize}
\item
  \emph{Check the validity of your set-up.} Having identified all
  relevant variables, you can operationalise the different dependent
  variable through different kinds of metrics. Here it is important to
  ensure the validity of your metrics. In other words: do your metrics
  measure what they are supposed to measure? One way to ensure the
  validity of a metric is to study the correlation between that metric
  and a trusted reference, such as human judgments. (Ehud \citet{reiter-2018-structured} investigated this in detail
  for the BLEU metric.) Alternatively you could think of theoretical
  arguments to motivate why your metric provides a good approximation of
  the variable of interest.
\item
  \emph{Select a sensible subset for evaluation.} If you cannot evaluate
  all the output of your model (and possibly the models that you are
  comparing your work to), think about the way you are sampling the
  outputs-to-be-evaluated. The sampling procedure heavily impacts the
  validity of your evaluation and the generalisability of your results.
\item
  \emph{Get IRB approval (if appropriate for your study).} When you know
  what the evaluation will look like, you can apply for approval with
  your local institutional review board (IRB, also known as `ethics
  committee') to determine that your study follows current ethics
  guidelines. If you do this at the onset of your project, you are less
  likely to run into any procedural delays. With a research proposal for
  the IRB in hand, you may also decide to turn the proposal into a full
  preregistration of your study (see
  \citealt{van-miltenburg-etal-2021-preregistering}).
\item
  \emph{Keep a log.} Carry out your study and note any deviations from
  your original plans, including the reasons why you changed your mind.
  These insights are essential to provide the rationale behind your
  study design. If you do not write this information down, you will
  forget it. For example, you could create a private GitHub repository
  to hold all your code, data, and notes. (Once your project is done,
  you can publish the repository alongside your paper. There are also
  \href{https://anonymous.4open.science}{services} to create a link to
  an anonymised version of your repository, that you can include in your
  submitted paper.) Or you could manage your project through the
  \href{https://osf.io}{Open Science Foundation} (OSF).
\item
  \emph{Be explicit about your materials and methods.} Report all
  relevant information about how you carried out your evaluation, so
  that others could reproduce your work using only your paper. If you
  carry out a human evaluation, the Human Evaluation DataSheet \citep{shimorina-belz-2022-human} provides an overview of important details to record.
\item
  \emph{Describe all relevant results}. If you are reporting overall
  scores, consider providing a table with disaggregated results for
  different subsets of the input data. Next to overview tables, you may
  also want to create insightful visualisations of the results. (Though
  choose wisely; don't just duplicate your results in another modality.)
  Go beyond the ``higher is better'' narrative and explain what the
  results mean for your system and the NLG literature in general. Be
  open about the limitations of your evaluation and the challenges that
  still lie ahead.
\end{enumerate}

\noindent \textbf{Bonus tip}: \emph{archive all data associated with your study.}
This includes including all outputs for the validation and test sets,
crowdsourcing templates, aggregated and
\href{https://pdai.info}{non-aggregated} human ratings, outputs of
statistical analysis software). Small files might be included in your
GitHub repository, but otherwise data can be hosted through other
services, e.g.~OSF, \href{https://zenodo.org}{Zenodo},
\href{https://figshare.com}{Figshare}, organisational repository or
national science hosting provider. It is possible that not all data can
be made available at submission time (though it is often possible to
share data anonymously), but at least try to be as exhaustive as
possible for your camera-ready version.

\section*{More to explore}
For more in-depth reading, here are some useful references:

\begin{itemize}
\item \citet{DBLP:journals/corr/abs-2006-14799} provide a survey of NLG evaluation methods.
\item \citet{gehrmann2022repairing} provide general recommendations regarding both human and automatic evaluations.
\item \Citet{VANDERLEE2021101151} provide recommendations for human evaluation
  studies.
\item \Citet{van-miltenburg-etal-2021-underreporting} provide guidelines for error analysis.
\item \citet{10.1145/3485766} provide an overview of automatic metrics used for NLG evaluation.
\item \href{https://ehudreiter.com/blog-index/}{Ehud Reiter's blog} has more
  recommendations for evaluating NLG systems.
\end{itemize}

If you have any questions, the community discord channel is a great
place to ask questions. Contact Dave Howcroft
(\href{mailto:D.Howcroft@napier.ac.uk}{\nolinkurl{D.Howcroft@napier.ac.uk}})
for an invitation.

\section*{Acknowledgments}
\emph{Thanks to Simone Balloccu, Ondřej Dušek, Dave Howcroft, Maria
Keet, Ember Manning, Maja Popović, Craig Thomson, and Sina Zarrieß for
feedback on an earlier draft of this essay.}


\bibliographystyle{acl_natbib}

\bibliography{sample.bib}


\end{document}